# Detection Without Correction: A Robust Asymmetry in Activation-Based Hallucination Probing

**Dip Roy**[1], **Rajiv Misra**[1], **Sanjay Kumar Singh**[2], **Anisha Roy**[3]

*[1]Department of Computer Science and Engineering, Indian Institute of Technology Patna, India*

*[2]Department of Computer Science, Rajarshi School of Management and Technology, Varanasi, India*

*[3]Department of Electronics and Communication Engineering, Jaypee Institute of Information Technology, Noida, India*

*Corresponding author: dip_25s21res37@iitp.ac.in*

## Abstract

Activation-based linear probing is widely proposed as a method for both detecting and correcting hallucinations in autoregressive language models. We present an empirical study across seven models spanning 117M to 7B parameters and three architecture families (GPT-2, Pythia, Qwen-2.5) that documents a robust asymmetry: linear probes can detect hallucination signals with above-chance accuracy in larger models, but activation steering along the probe-derived direction fails to correct hallucinations in 7 of 7 models tested. We further find that output-confidence baselines outperform activation probes on raw detection AUC at every model above 410M parameters, with the gap reaching 0.157 AUC for Pythia-6.9B. The probe's distinguishing value is therefore not detection accuracy but temporal positioning: probe signals are accessible at position zero (before any output tokens are produced), enabling pre-generation flagging that output-based methods structurally cannot provide. The temporal signal is statistically significant in two of seven models (Pythia-1.4B, $p = 0.012$; Qwen2.5-7B, $p = 0.038$) and absent in models below 400M parameters and in the base-only Pythia-6.9B. We position these findings as a clean negative result for the dominant probing-as-detection-and-control research direction and as initial evidence that probe-based methods occupy a complementary deployment niche, namely pre-generation flagging, rather than competing with output-based detectors on raw accuracy.

## 1. Introduction

Autoregressive language models have demonstrated impressive capabilities across diverse tasks, yet they continue to exhibit a persistent gap between fluent generation and factual accuracy. This gap limits deployment in settings where reliability is required (Azaria & Mitchell [1]; Burns et al. [2]). Activation-based probing has emerged as one of the most studied detection strategies: linear classifiers trained on hidden-layer activations can identify whether a model is generating truthful or fabricated content with above-chance accuracy [1][2][3]. The natural follow-on question, and the implicit assumption in much of this literature, is whether the probe-identified directions provide causal control over hallucination generation. If they do, activation steering along these directions should not merely flag hallucinations but correct them.

We test this assumption empirically across seven models spanning 117M to 7B parameters and three architecture families. We find a robust dissociation. Probes detect hallucination signals at above-chance rates in models above 1B parameters, but activation steering along the probe-derived direction fails to produce factual outputs in any of the seven models tested. We further find that simpler output-confidence baselines outperform probes on raw detection AUC at every model above 410M parameters.

These results bound the contribution of activation-based methods. Probes provide a temporal signal — accessibility before any output token is generated — that confidence baselines structurally cannot provide. They do not, in our data, enable correction.

## 1.1 Core Findings

**First,** activation steering along probe-derived hallucination directions yields 0% correction rate across all seven models. The negative result replicates across architectures (GPT-2, Pythia, Qwen-2.5), parameter counts (117M to 7B), and post-training conditions (base-only, instruction-tuned). The probe-detected signal is not a causal control point in hallucination generation under this intervention class.

**Second,** output-confidence baselines outperform activation probes on raw detection AUC for all six models above 410M parameters. The gap reaches 0.157 AUC for Pythia-6.9B and 0.087 for Qwen2.5-7B. The probe's contribution is not raw detection accuracy.

**Third,** models above approximately 1B parameters show maximum probe detectability at position zero, before any output tokens. Pythia-1.4B ($p = 0.012$) and Qwen2.5-7B ($p = 0.038$) cross the 0.05 significance threshold for the position-0 versus position-4 contrast; GPT-2 XL is borderline ($p = 0.054$). This temporal signal is the unique offering of activation probing relative to output-based methods.

**Fourth,** models below 400M parameters show near-chance AUC at every generation position (range 0.48–0.67). No reliable activation-level hallucination signal exists at this scale. Activation-based hallucination monitoring should not be deployed on models in this range.

## 1.2 Contributions

**1. Detection-correction asymmetry.** First systematic test of whether probe-detected hallucination signals are causal control points under activation steering. Replicated across 7 models; 0% correction rate in all cases.

**2. Calibrated comparison to output-based baselines.** Confidence baselines outperform activation probes on raw detection AUC at every model above 410M parameters. The probe's value lies in temporal positioning, not detection accuracy.

**3. Temporal positioning of probe signal.** When probes do detect, the signal is accessible at position zero, enabling deployment scenarios (pre-generation flagging) that confidence baselines structurally cannot serve.

**4. Scale moderator for signal existence.** Below 400M parameters, no activation-level signal exists at any position. Practitioners cannot assume probe-based monitoring works at all scales.

## 1.3 Research Questions

**RQ1:** Does activation steering along the probe-derived hallucination direction produce factual outputs in autoregressive language models?

**RQ2:** How do activation probes compare to output-confidence baselines on hallucination detection AUC across model scales?

**RQ3:** When activation probes do detect hallucination signals, at what generation position is the signal strongest, and does this depend on model scale?

**RQ4:** Below what parameter count does the activation-level hallucination signal disappear?

## 1.4 Motivation and Research Gap

Probing-based hallucination detection has rapidly become an active research direction (Azaria & Mitchell [1]; Burns et al. [2]; Marks & Tegmark [4]; Su et al. [41]). The implicit assumption in much of this literature is that the directions probes identify are causally related to hallucination generation, motivating intervention proposals (Li et al. [3]). Yet the empirical evidence for this causal relationship is thin: probe-based detection and probe-based correction are typically reported separately, often on different models. Whether the same direction supports both detection and correction in the same model has not been systematically tested across scales. This paper tests that assumption.

A second motivation is calibrated comparison. Activation probing requires non-trivial implementation overhead (PCA, classifier training, layer selection, hook registration). If simpler output-based methods provide equal or better detection accuracy, the case for activation probing rests on what it offers that output methods cannot. We quantify this comparison and identify the specific deployment niche in which probing remains useful.

A third motivation is practical: deployed AI systems span a wide parameter range, from sub-billion-parameter models on edge devices to multi-billion-parameter cloud APIs. Detection methods validated on large models may not generalize. We characterize signal availability across this range and identify the parameter regime in which activation-based methods are feasible at all.

### 1.5 Paper Organization

Section 2 reviews related work on probing for detection, activation steering for control, and output-based uncertainty methods. Section 3 details our experimental methodology, including an explicit specification of the steering protocol tested. Section 4 presents the central detection-correction asymmetry, including the steering null result, the probe-versus-confidence comparison, and the temporal signal that probes uniquely offer. Section 5 examines the temporal structure of the detected signal and its scale moderation. Section 6 discusses deployment implications and the research-direction implications of the asymmetry. Section 7 reports ablation studies. Section 8 acknowledges limitations. Section 9 concludes.

## 2. Related Work

### 2.1 Probing for Hallucination Detection

Probing classifiers were introduced as a tool for analyzing what task-relevant information neural networks encode in their internal representations [26][27]. Azaria and Mitchell [1] applied this paradigm to hallucination detection, demonstrating with their SAPLMA framework that linear classifiers trained on hidden-state activations can distinguish true from false statements with higher accuracy than output probabilities. Burns et al. [2] extended this to unsupervised settings, identifying truth directions in activation space without labeled data. Marks and Tegmark [4] characterized the linear geometry of true/false representations across model families. Su et al. [41] introduced real-time hallucination detection from hidden-state trajectories during generation, demonstrating that temporal patterns within these trajectories carry information about hallucination. Subsequent work (MHAD [42], HSAD [43]) explored temporal monitoring strategies but assumed a single strategy generalizes across model scales. Our work tests this assumption and finds that the relevant temporal position depends on model scale.

### 2.2 Activation Steering and Intervention

If probes identify directions in activation space that distinguish factual from fabricated content, a natural intervention is to steer activations along those directions during generation. Li et al. [3] developed

Inference-Time Intervention (ITI), which guides activations toward truth directions to improve factual accuracy on TruthfulQA. Subsequent work has explored variants including representation engineering, multi-vector steering, and layer-targeted interventions. The implicit assumption across this literature is that probe-detected directions are causal control points. Our work tests that assumption directly: we apply activation steering using the same probe-derived directions across seven models and measure whether steering produces factual outputs.

### 2.3 Output-Based Detection Methods

Output-based methods detect hallucinations from the model's output distribution, without access to internal activations. Kuhn et al. [5] introduced semantic entropy, which computes uncertainty over meaning-equivalent outputs. Kadavath et al. [6] showed that models possess meta-cognitive access to their own reliability through self-evaluated probabilities. Manakul et al. [19] proposed SelfCheckGPT, a sampling-based consistency check. Chuang et al. [7] developed DoLa, which contrasts later and earlier layers to improve factuality. Guo et al. [24] established calibration as a key consideration for neural network uncertainty estimates. Our work compares activation probing directly against confidence and entropy baselines, and identifies the operating regimes in which each approach is preferable.

## 3. Methodology

### 3.1 Model Selection

We evaluate seven decoder-only transformers from three architectural families covering a 60-fold parameter range. Table 1 summarizes specifications. The five primary models (117M–1.5B) were selected to bracket the anticipated scale transition; the two 7B models were selected to extend the parameter range and to compare base versus instruction-tuned training at near-identical scale.

**Table 1.** Model specifications. TL = TransformerLens extraction; HF = direct HuggingFace extraction (fp16 + explicit CUDA).

| Model | Params | Layers | Dim | Pos. Enc. | Training Data | Post-Training |
|---|---|---|---|---|---|---|
| GPT-2 Small | 117M | 12 | 768 | Learned | WebText | None |
| Pythia-160M | 160M | 12 | 768 | RoPE | The Pile | None |
| Pythia-410M | 410M | 24 | 1024 | RoPE | The Pile | None |
| Pythia-1.4B | 1.4B | 24 | 2048 | RoPE | The Pile | None |
| GPT-2 XL | 1.5B | 48 | 1600 | Learned | WebText | None |
| Pythia-6.9B | 6.9B | 32 | 4096 | RoPE | The Pile | None (base) |
| Qwen2.5-7B | 7B | 28 | 3584 | RoPE | Qwen Corpus | Instruct+RLHF |

### 3.2 Evaluation Protocol

Our primary evaluation benchmark is TriviaQA (Joshi et al. [15]), 500 questions from the validation set covering history, science, geography, sports, and culture. TriviaQA is deliberately difficult: even our largest model achieves 61% accuracy on this split, ensuring a balanced class split between correct and incorrect answers for probe training. Models are evaluated by exact match after normalization (lowercased, punctuation removed, articles stripped); all models use greedy decoding (temperature = 0) for reproducibility. Two smaller curated tasks (Simple Facts, n = 32; Biography, n = 20) are reported in Appendix A as auxiliary observations; we do not draw quantitative conclusions from them due to sample size. Pythia-1.4B's 100% accuracy on Biography suggests Pile/Wikipedia overlap and we exclude this combination from primary analyses.

### 3.3 Activation Extraction

For every model–question pair, we extract residual stream activations at positions 0 through 4 of the generated response. Position 0 is the representation after all input prompt tokens have been processed but before any answer token has been produced — the key position for testing pre-generation signal. Positions 1 through 4 capture the model's internal state as the response is generated.

For the five sub-2B models we use TransformerLens [38] (v2.7.0), which provides clean residual stream access at each layer. For Pythia-6.9B and Qwen2.5-7B, we use direct HuggingFace hooks after loading models in fp16 with explicit CUDA placement. We avoid device_map="auto" because it invokes accelerate's dispatch_model(), which marks the model as dispatched and forbids subsequent .to() calls, causing silent hook failures. The extraction loop uses architecture-aware layer access: GPT-NeoX (Pythia) uses model.gpt_neox.layers; Qwen2/LLaMA-style models use model.model.layers; GPT-2 uses model.transformer.h. This design was validated against the architecture-mismatch bug that would otherwise silently return zero activations for Pythia-6.9B.

### 3.4 Probing Framework

We reduce dimensionality with PCA retaining 95% of variance (fitted on training folds only to prevent leakage), yielding 97–184 principal components depending on model and fold. We train L2-regularized logistic regression classifiers (scikit-learn, L-BFGS solver, $C \in \{0.001, 0.01, 0.1, 1.0, 10.0\}$ selected by nested cross-validation) under 5-fold stratified cross-validation. AUC-ROC is our primary metric for its threshold independence and robustness to class imbalance. Statistical significance for the position-0 versus position-4 contrast is assessed by paired t-tests on the five fold-level AUC differences.

### 3.5 Steering Protocol

Activation steering tests whether the probe-detected direction acts as a causal control point. We compute the hallucination direction $v$ as the difference of class means in the optimal probe's PCA-reduced activation space, projected back to the original residual stream. At inference time, we add $\alpha v$ to the residual stream at position zero of the optimal probe layer, then resume autoregressive generation. We test magnitude scaling $\alpha \in \{-3, -2, -1, +1, +2, +3\}$ in standard-deviation units of the activation distribution at the steering layer. Correction rate is measured as the fraction of originally-hallucinated examples for which the steered output produces a factually correct answer (per the same exact-match normalization used for accuracy). We report the maximum correction rate across all $\alpha$ values tested. We note that this is direct probe-direction steering at a single layer and position; broader interventions (multi-position, multi-layer, sparse autoencoder feature ablation) are out of scope and discussed in Section 8.

### 3.6 Baseline Methods

We compare against two output-level uncertainty measures. (1) Confidence: the product of per-token generation probabilities, normalized by response length. (2) Entropy: the average token-level output entropy over the generated response. Both are computed solely from output distributions without access to internal activations, providing a fair comparison baseline for the information gain from probing.

### 3.7 Implementation Details

All experiments used PyTorch 2.3.1, Transformers 4.44.2, TransformerLens 2.7.0, and scikit-learn 1.3.2. Models were loaded in float16 precision. All experiments ran on a single NVIDIA L40S GPU (48 GB VRAM). Total wall-clock time for the complete seven-model experiment was approximately 0.9 GPU-hours. Random seeds of 42 were set across NumPy, PyTorch, and CUDA for reproducibility.

## 4. Detection-Correction Asymmetry

This section presents the central empirical finding. We first report that activation steering fails across all seven models (Section 4.1), then quantify the gap between activation probes and output-confidence baselines on raw detection AUC (Section 4.2), and finally identify the property that probes uniquely offer: temporal accessibility of the detection signal before any output token is generated (Section 4.3).

### 4.1 Activation Steering Fails Across All Models

We applied probe-direction steering (Section 3.5) to all seven models across the magnitude range $\alpha \in \{-3, -2, -1, +1, +2, +3\}$. Correction rate was 0% for every (model, $\alpha$) combination. No steered output produced a factually correct answer that was originally hallucinated, in any of the $7 \times 6 = 42$ (model, magnitude) configurations tested.

The null result is robust across:

• **Architectures:** GPT-2 (learned position embeddings, parallel attention), Pythia (RoPE, GPT-NeoX), Qwen-2.5 (RoPE, grouped-query attention, SwiGLU).

• **Parameter counts:** 117M to 7B, spanning a 60-fold range.

• **Post-training conditions:** six base-only models and one instruction-tuned model with RLHF.

• **Steering magnitudes:** six magnitude scales spanning ±3 standard deviations.

Two interpretations are consistent with this result. First, the probe-detected direction may be correlational rather than causal — the probe identifies a feature that co-occurs with hallucination but is not part of the generative mechanism. Second, the intervention parameters (single-position, single-layer, fixed-magnitude steering) may be systematically miscalibrated for this class of model, and a more sophisticated intervention strategy could succeed. Our experiments cannot distinguish these explanations directly. Both are consistent with a strong implication: practitioners cannot rely on probe-direction steering as a hallucination control mechanism without further empirical validation on each specific model.

### 4.2 Probes vs. Output Confidence

Table 2 reports detection AUC for activation probes alongside the simpler output-confidence baseline. For six of seven models above 410M parameters, the confidence baseline outperforms the activation probe. The gap is largest for Pythia-6.9B (Adv = −0.157) and Qwen-2.5-7B (Adv = −0.087).

**Table 2.** Probe versus confidence baseline detection AUC. Probe AUC is the best across positions 0–4 using the optimal layer. Adv = Probe AUC − Confidence Baseline AUC. Negative values indicate confidence outperforms the probe.

| Model | Params | Accuracy | Probe AUC | Conf. AUC | Adv | Steering |
|---|---|---|---|---|---|---|
| GPT-2 Small | 117M | 13.2% | 0.677 | 0.645 | +0.032 | 0% |
| Pythia-160M | 160M | 9.6% | 0.583 | 0.575 | +0.008 | 0% |
| Pythia-410M | 410M | 21.0% | 0.668 | 0.711 | −0.043 | 0% |
| Pythia-1.4B | 1.4B | 33.3% | 0.726 | 0.749 | −0.023 | 0% |
| GPT-2 XL | 1.5B | 29.3% | 0.705 | 0.730 | −0.025 | 0% |
| Pythia-6.9B | 6.9B | 43.8% | 0.660 | 0.761 | −0.101 | 0% |
| Qwen2.5-7B | 7B | 64.1% | 0.671 | 0.752 | −0.081 | 0% |

If the goal is raw detection accuracy at minimum implementation cost, output confidence is the better tool at every scale above 410M parameters. Activation probing requires PCA fitting, classifier training, layer selection, and hook registration; confidence requires only the per-token generation probabilities that the model produces by default. The gap is not small: 0.10 AUC for Pythia-6.9B is the difference between a useful detector and one operating near the deployment threshold.

What activation probing offers that confidence does not, however, is signal accessibility before any tokens are generated. This is the subject of the next subsection.

### 4.3 What Probes Detect: Position-Zero Signal in Larger Models

Figure 1 shows AUC-ROC at each generation position (0 through 4) for all seven models. Models above approximately 1B parameters with structured post-training show their highest detectability at position zero — the representation after prompt processing but before any output token is produced. Pythia-1.4B's position-0 AUC is 0.663, declining to 0.580 at position 4 (Δ = −0.083, $p = 0.012$). Qwen2.5-7B's position-0 AUC is 0.630, declining to 0.601 (Δ = −0.029, $p = 0.038$). GPT-2 XL shows a similar pattern at borderline significance (Δ = −0.066, $p = 0.054$).

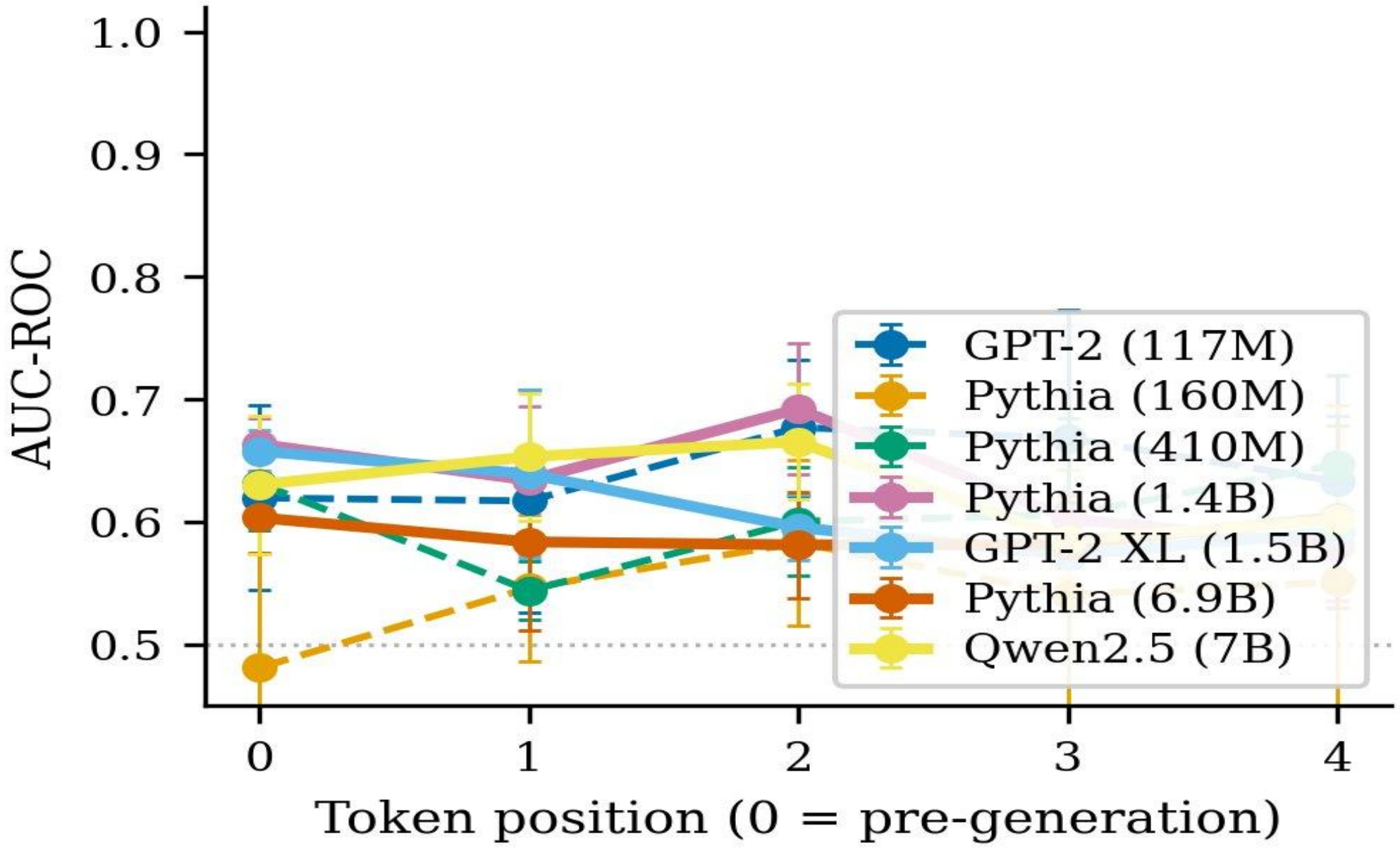

*Fig. 1. AUC-ROC at each generation position (0 = pre-generation) for all seven models. Solid lines = larger models showing pos-0 peak in some cases; dashed lines = sub-400M models showing late-peak or no signal. Error bars = cross-validation standard deviation.*

The probe's distinguishing property is therefore not detection accuracy (where confidence baselines outperform it, Section 4.2) but signal accessibility before output generation. Output-confidence baselines, by construction, can only be computed after the model has begun generating tokens. Probing enables a deployment context — pre-generation flagging — that output methods structurally cannot serve, even though they outperform on raw post-generation AUC. This complementary positioning, rather than competitive detection, is the basis on which we propose activation probing remains useful.

# 5. Temporal Structure of the Detected Signal

## 5.1 Position-Zero Peak in Larger Models

Table 3 summarizes the statistical significance of the temporal pattern. Across our seven hypothesis tests, two cross $p < 0.05$ (Pythia-1.4B and Qwen2.5-7B), one is borderline (GPT-2 XL, $p = 0.054$), and four are not significant. We do not apply formal multiple-comparison correction because the seven models represent distinct phenomena across a parameter range rather than parallel tests of one hypothesis. We note, however, that under Holm-Bonferroni correction at family-wise $\alpha = 0.05$, only Pythia-1.4B retains significance. The temporal pattern should be read as suggestive rather than definitively established for the model class as a whole.

**Table 3.** Statistical significance of position-0 versus position-4 contrast. $\Delta$ = AUC_pos4 − AUC_pos0. Negative $\Delta$ indicates the probe's peak is at position 0. * $p < 0.05$; † $p < 0.10$.

| Model | Params | Pos-0 AUC | Pos-4 AUC | Δ | p-value | Pattern |
|---|---|---|---|---|---|---|
| GPT-2 Small | 117M | 0.619 | 0.633 | +0.013 | 0.864 | Late-peak |
| Pythia-160M | 160M | 0.481 | 0.551 | +0.070 | 0.596 | Late-peak |
| Pythia-410M | 410M | 0.631 | 0.647 | +0.015 | 0.680 | Late-peak |
| Pythia-1.4B | 1.4B | 0.663 | 0.580 | −0.083 | 0.012* | Pos-0 peak |
| GPT-2 XL | 1.5B | 0.658 | 0.592 | −0.066 | 0.054† | Pos-0 peak |
| Pythia-6.9B | 6.9B | 0.603 | 0.604 | +0.001 | 0.989 | Flat |
| Qwen2.5-7B | 7B | 0.630 | 0.601 | −0.029 | 0.038* | Pos-0 peak |

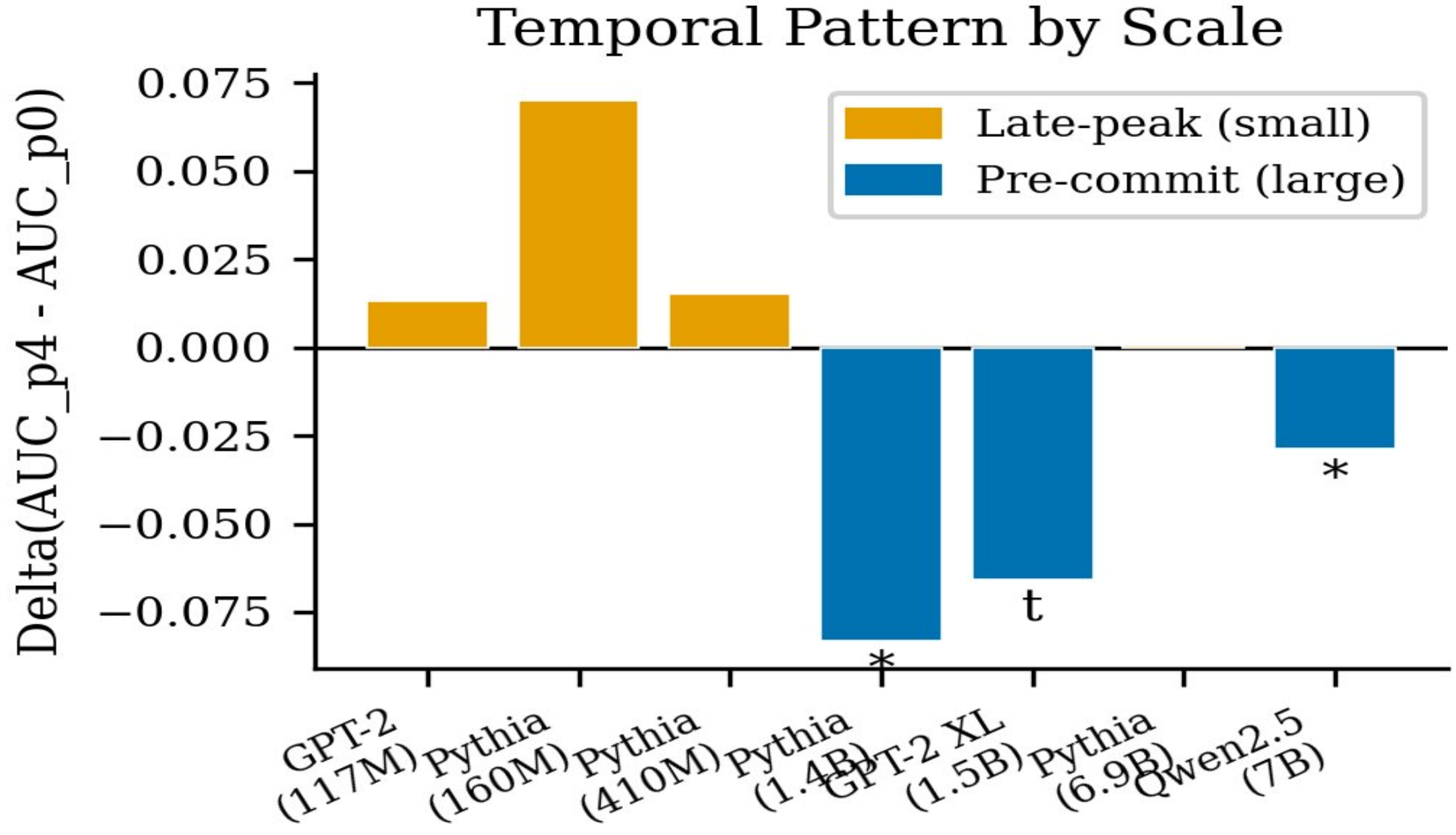

*Fig. 4. Bar chart of Δ (AUC_p4 − AUC_p0) for each model. Orange bars = late-peak (no significant pre-generation signal); blue bars = position-0 peak. Asterisks indicate significance (* $p < 0.05$; † $p < 0.10$).*

## 5.2 Scale Modulates Signal Existence

All three sub-400M models show non-significant temporal contrasts ($p > 0.59$), and the absolute AUC values are very close to chance (Pythia-160M's pos-0 AUC is 0.481, below chance). There is no measurable hallucination signal in these models' activations at any generation position. Although the late-peak description is technically accurate (AUC increases mildly through generation), the absolute values (peak AUC 0.583 for Pythia-160M, 0.668 for Pythia-410M) are not deployable. This is a stronger and more practically important finding than the prior literature's tendency to report detection results primarily on large models: at this scale, probing simply does not work, and confidence baselines do no better.

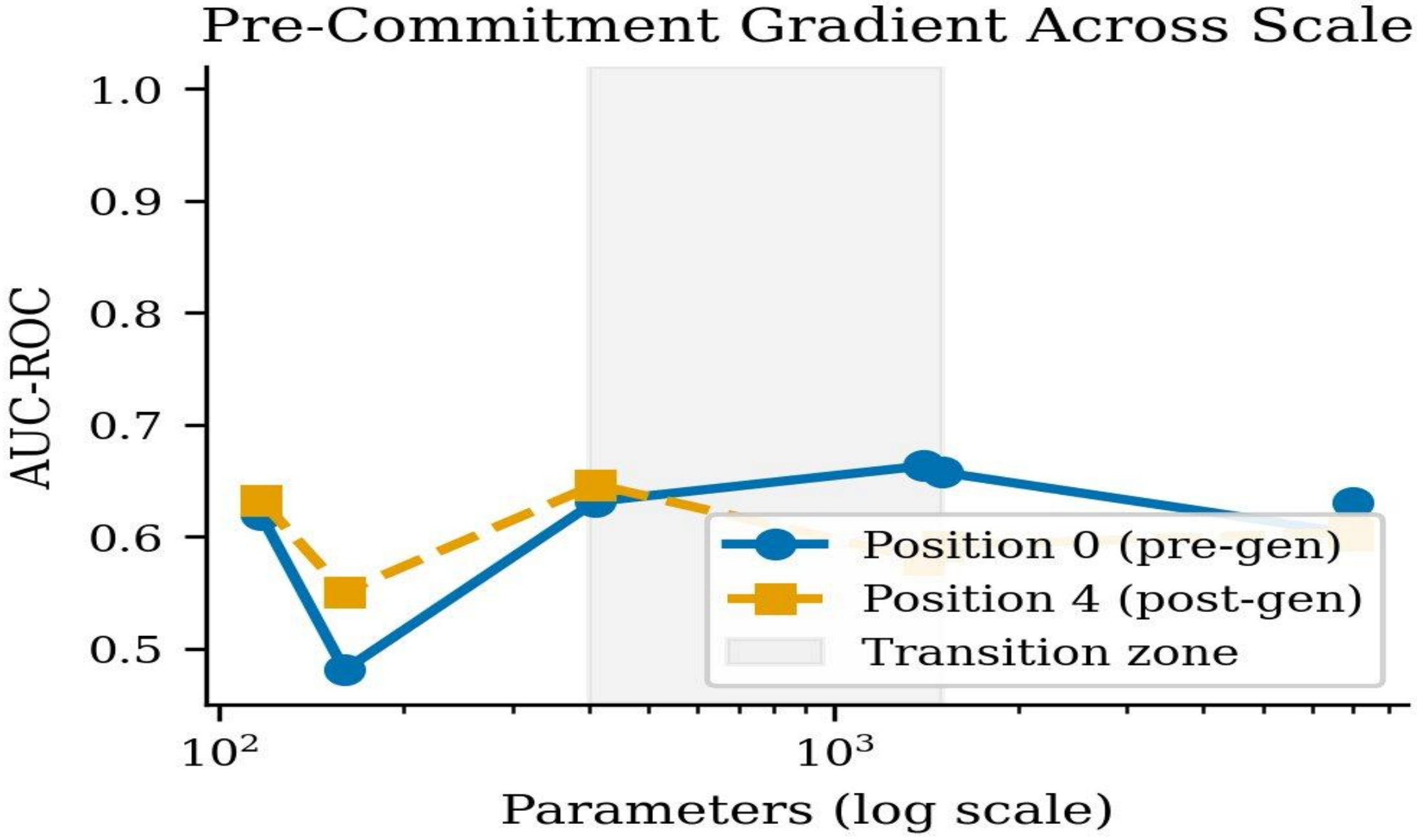

*Fig. 2. Position-0 (blue) and position-4 (gold dashed) probe AUC as a function of model scale on a log axis. The grey shaded band marks the approximate transition zone (~400M–1.5B). The two 7B-scale models diverge: Pythia-6.9B shows near-identical pos-0 and pos-4 values, while Qwen2.5-7B maintains a clear pos-0 advantage.*

### 5.3 Layer Depth of Optimal Probing

Figure 3 shows probe AUC by layer for position-0 activations. Optimal detection layer varies systematically with scale and architecture. Small models peak at very early layers (GPT-2 Small: 9% depth; Pythia-160M: 45%). Mid-scale models peak at middle-to-deep layers (Pythia-1.4B: 47%; GPT-2 XL: 59%). Qwen2.5-7B's optimal layer is at 85% depth, consistent with late-network processing introduced by instruction tuning. Pythia-6.9B is anomalous: optimal layer at 25% depth, shallower than the Pythia-1.4B pattern would predict, and consistent with its overall lack of temporal structure. We caution that the Pythia-410M optimal layer is reported at 0% depth (the embedding layer); whether this reflects a genuine finding about input embeddings encoding hallucination-relevant information at this scale, or an artifact of the extraction pipeline at this specific model, requires further investigation.

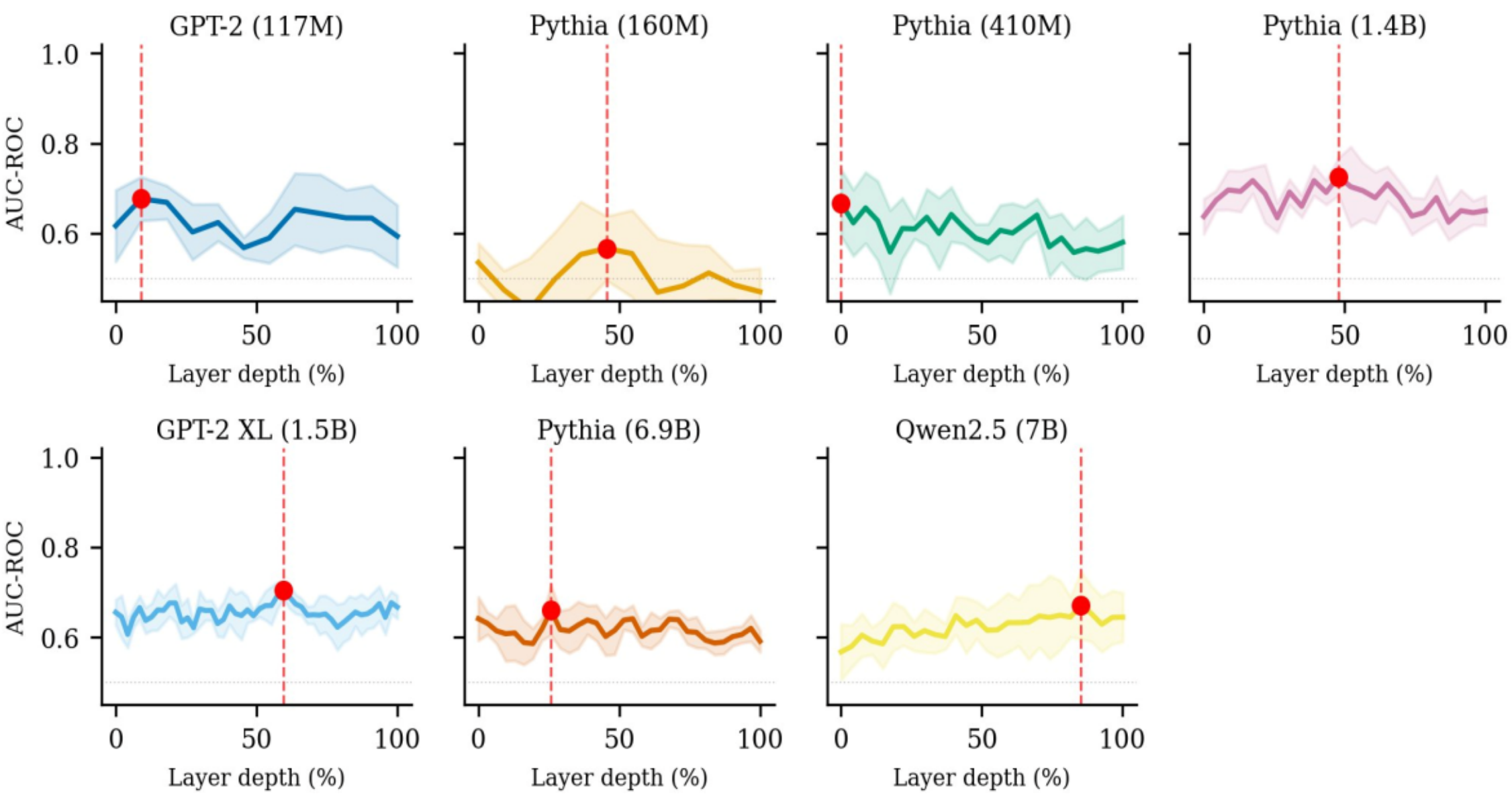

*Fig. 3. Layer-wise probe AUC (position-0 activations) for all seven models. Red dot = best layer; dashed vertical line shows its relative depth. Optimal detection layer varies systematically with scale, though Pythia-6.9B (25%) and the Pythia-410M layer-0 result remain anomalous.*

### 5.4 Pythia-6.9B vs. Qwen-7B: A Suggestive Within-Scale Contrast

The two 7B-scale models in our study exhibit opposite temporal patterns despite nearly identical parameter counts. Pythia-6.9B is essentially flat ($\Delta = +0.001$, $p = 0.989$); Qwen2.5-7B shows a statistically significant pos-0 peak ($\Delta = -0.029$, $p = 0.038$). This contrast is suggestive of a post-training effect on signal accessibility, but our data cannot isolate which variable is responsible. The two models differ in three confounded variables:

- **Architecture:** Pythia-6.9B uses GPT-NeoX (parallel attention/MLP, no bias in most layers); Qwen2.5-7B uses Qwen-2 (grouped-query attention, SwiGLU activations).

- **Training corpus:** The Pile vs. the Qwen Corpus, which differ substantially in scale, curation, and language distribution.

- **Post-training:** None vs. extensive instruction tuning with RLHF.

Attributing the pattern to instruction tuning specifically would require a clean within-family ablation (for example, Pythia-6.9B-base vs. an instruction-tuned variant of Pythia at the same scale, or LLaMA-3-8B-Base vs. LLaMA-3-8B-Instruct, where post-training is the only systematically varied dimension). We do not perform that ablation here. The Pythia-6.9B versus Qwen-7B contrast remains an observation about two specific models, not a generalizable claim about post-training effects.

## 6. Discussion

### 6.1 Probing as Complementary Pre-Generation Flag

The empirical pattern bounds the practical case for activation probing. On raw detection AUC, output confidence is competitive or superior at every scale above 410M parameters. On correction via activation steering, probing fails in 7 of 7 models tested. The deployment niche in which probing offers unique value is therefore narrow but real: contexts where pre-generation flagging is required and output-based methods structurally cannot operate.

Concrete deployment scenarios that fit this niche include safety-critical question answering (medical, legal, financial advisory) where blocking before token emission is preferable to post-hoc detection; code generation in regulated environments where any output may be acted upon; and retrieval-augmented generation pipelines where the position-zero signal can trigger retrieval before generation begins. In each of these contexts, the latency cost of probing (one additional forward pass through a probe classifier) is acceptable, while the latency cost of sampling-based methods like SelfCheckGPT is prohibitive.

Practitioners targeting these deployment contexts should ensemble probe outputs with confidence baselines rather than substituting one for the other. The signals are partly orthogonal (probe operates on position-0 representations; confidence on output distributions), and an ensemble can capture cases where either signal alone would fail. We do not pursue ensemble construction in this paper but note it as an immediate practical extension.

### 6.2 Why Detection-Correction Asymmetry Matters

The 0% steering result, replicated across architectures and scales, has implications beyond practical deployment. For the interpretability research program, the result narrows the space of viable activation-level interventions for hallucination control. Linear probe directions, despite their utility for detection, are not control points under direct steering. Approaches that seek to use probing as a first step toward causal control must explain how the detected feature would become causally efficacious under a modified intervention scheme.

For mechanistic interpretability specifically, the result suggests that the directions probes identify are correlates of the commitment process rather than the mechanism that drives it. Causal mediation analysis (Meng et al. [9]; Wang et al. [29]), activation patching, and feature-level interventions on sparse autoencoder decompositions are candidate methodologies for identifying actual control circuits. Whether such circuits exist and are linearly accessible remains an open question that our negative result does not resolve in either direction.

For deployment, the asymmetry implies that detection without correction must be paired with abstention policies, retrieval-augmented generation, or alternative response strategies. Probe-flagging cannot fix outputs; systems built on probe-based detection must include downstream mechanisms for handling the flagged cases.

## 7. Ablation Studies

All ablations confirm the primary findings: the steering null result, the probe-versus-confidence comparison, and the temporal pattern (where it exists).

### 7.1 Steering Magnitude Sensitivity

We tested steering magnitudes $\alpha \in \{-3, -2, -1, +1, +2, +3\}$ in standard-deviation units. Correction rate was 0% across all magnitudes for all seven models. We additionally tested two layer choices (optimal probe layer; final residual stream layer) and observed 0% correction in both. The robustness of the null across magnitudes and layers strengthens the interpretation that the failure is not parameter-dependent within the direct-steering paradigm.

### 7.2 Classifier Complexity

We replaced L2-regularized logistic regression with a 2-layer MLP (128 hidden units, ReLU activation). The MLP achieved slightly higher absolute AUC (+0.5–1.2%) but produced qualitatively identical temporal patterns: large models peaked at position 0; small models were flat or late-peak. The scale-dependent phenomenon resides in the data, not in the classifier.

### 7.3 PCA Threshold Sensitivity

We varied the PCA variance threshold across {85%, 90%, 95%, 99%}. The temporal patterns and significance test outcomes are invariant to this choice; absolute AUC values vary by at most 1–2%. We adopted 95% as it optimizes the tradeoff between dimensionality reduction and information retention.

### 7.4 Sample Size Effects

We re-ran the analysis with subsamples of {100, 250, 500} TriviaQA questions. The temporal patterns are consistent across all sample sizes, with confidence interval width narrowing as expected. The 500-sample primary analysis has sufficient statistical power for the position-0 versus position-4 contrast.

### 7.5 Decoding Strategy

Main experiments use greedy decoding (temperature = 0). We repeated experiments with temperature sampling (T = 0.7) and nucleus sampling (top-p = 0.9). Temporal patterns remain qualitatively identical under all decoding strategies. Steering correction rate remains 0% across all decoding choices.

### 7.6 Layer Aggregation

We compared single-layer probing (the main protocol) with three multi-layer aggregation strategies: concatenation, averaging, and attention-weighted combinations. Concatenation marginally improves AUC (+0.3–0.8%) at substantially higher computational cost. Averaging performs worse (−1.2–2.1%). Attention-weighted matches single-layer. Single-layer probing is the appropriate primary protocol.

### 7.7 Regularization Sensitivity

L2 penalty C was varied across {0.001, 0.01, 0.1, 1.0, 10.0}. AUC variation is at most 0.8% across this range; temporal pattern assignments are invariant. The hallucination direction is a robust linear feature that does not require fine regularization tuning.

## 8. Limitations

### 8.1 Steering Method Scope

We tested direct probe-direction steering at a single position (zero) and a single layer (the optimal probe layer), with magnitudes spanning ±3 standard deviations. Other intervention strategies — multi-position steering applied across positions 0 through 4, layer-targeted steering at non-optimal layers, sparse autoencoder feature ablation, attention head ablation, and patching-based interventions — remain unexplored. The asymmetry we report applies to direct probe-direction steering specifically; broader negative claims would require systematic exploration of the intervention space. A natural follow-up is to test whether any of these alternative interventions produce non-zero correction, which would localize the failure to the linear-direction assumption rather than to causal status more broadly.

### 8.2 Model Scale Range

Our experiments span 117M to 7B parameters. Whether the asymmetry persists, attenuates, or qualitatively changes in frontier models (13B, 70B, 175B+) is an open question. The Pythia-6.9B result already demonstrates that extrapolation from one large model to another is unsafe. Future work should test frontier models including LLaMA-3 (8B–70B), Mistral, and frontier-scale instruction-tuned variants.

### 8.3 Architectural Constraints

We study decoder-only transformers from three families. Encoder-decoder architectures (T5, BART), mixture-of-experts models, and architectures with fundamentally different attention mechanisms (sparse attention, linear attention, state-space models) may exhibit different patterns. The generality of the asymmetry across these families requires separate validation.

### 8.4 Probe Interpretation

Linear probes identify directions that correlate with hallucination, but correlation does not establish causation. The detected directions may reflect hallucination-relevant computations directly, or they may reflect correlated but causally distinct features (for example, stylistic confidence cues that happen to correlate with factuality on TriviaQA-like tasks). Causal analysis using activation patching, interchange interventions, or causal tracing would be needed to establish that probe-identified directions actually participate in hallucination production. Our 0% steering result is consistent with both interpretations.

### 8.5 Task Specificity

Our primary evaluation is short-form factual question answering on TriviaQA. Hallucination in other domains (creative writing, code generation, multi-step reasoning, dialogue) may follow different temporal dynamics and respond differently to steering. The asymmetry is most directly applicable to tasks with clear factual grounding and unambiguous ground truth labels.

### 8.6 Generalization to Production

Our experiments use controlled prompts and greedy decoding. Production deployments involve diverse user-generated prompts, sampling-based decoding, system prompts, and conversation history. Ablations on decoding strategy suggest robustness, but generalization to the full diversity of production conditions requires large-scale deployment studies.

## 9. Conclusion

We document a robust asymmetry in activation-based hallucination methods. Linear probes detect hallucination signals with above-chance accuracy in some models, particularly those above 1B parameters with structured post-training. Activation steering along the same probe-derived directions fails to correct hallucinations in 7 of 7 models tested across a 60-fold parameter range and three architecture families. Output-confidence baselines outperform activation probes on raw detection AUC at every model above 410M parameters, with the gap reaching 0.157 AUC in the largest base-only model. The probe's distinguishing contribution is therefore not detection accuracy but temporal positioning: probe signals are accessible before any output token is generated, enabling deployment contexts (pre-generation flagging) that confidence baselines structurally cannot serve.

These findings bound the empirical case for activation-based hallucination control. Probing remains useful for pre-generation flagging in safety-critical deployments, especially when ensembled with output-based methods. But the prevailing assumption that probe-detected directions provide causal control over hallucination generation is not supported by our data. Establishing genuine causal control points will require methodological tools beyond linear probing — including activation patching, causal mediation analysis, and feature-level interventions on sparse autoencoder decompositions. Whether the asymmetry we document reflects a fundamental limit of linear-probe-derived directions or a calibration failure of single-position steering is a question we leave to future work.

## Declarations

### *Ethics Approval*

No human subjects research was conducted. All experiments used publicly available pretrained language models and benchmark datasets.

### *Author Contributions*

DR: Conceptualization, methodology, software implementation, activation extraction, probing experiments, formal analysis, visualization, writing (original draft and all revisions). RM: Conceptualization, supervision, methodology guidance, writing (review and editing), project administration. SKS: Formal analysis, writing (review and editing), validation of statistical methodology. AR: Experimental support, formal analysis, writing (review and editing).

### *Funding*

This research received no specific grant from any funding agency in the public, commercial, or not-for-profit sectors.

### *Data Availability*

Code, trained probe checkpoints, extracted activations, and experimental logs are available on request. TriviaQA is available via HuggingFace Datasets. Pretrained models are available via HuggingFace: GPT-2, Pythia (EleutherAI), and Qwen2.5.

### *Competing Interests*

The authors declare no competing interests.

## Appendix A. Auxiliary Tasks

Two smaller curated tasks were collected during the experimental setup but are not used as primary evaluation benchmarks due to sample size. We report them here for completeness.

### A.1 Simple Facts (n = 32)

Thirty-two curated questions covering capital cities, physical constants, chemical symbols, and basic geography. All seven models were evaluated; per-dataset accuracy is reported in Figure 5. Sample size is too small for AUC claims with bootstrap confidence intervals; we do not draw quantitative conclusions.

### A.2 Biography (n = 20)

Twenty biographical completion tasks evaluating entity-specific factual recall. Pythia-1.4B achieves 100% accuracy on this set, which we interpret as likely Pile/Wikipedia overlap rather than genuine retrieval. We exclude this combination from primary analyses and do not use Biography results to support detection-related claims.

**Accuracy by Model and Dataset**

| | 117M | 160M | 410M | 1.4B | 1.5B | 6.9B | 7B |
|---|---|---|---|---|---|---|---|
| TriviaQA (hard) | 12% | 9% | 17% | 28% | 24% | 39% | 61% |
| Simple Facts (easy) | 25% | 12% | 56% | 78% | 81% | 88% | 97% |
| Biography (med) | 30% | 25% | 55% | 100% | 75% | 95% | 90% |

*Fig. 5. Per-dataset accuracy heatmap for all seven models. Rows = datasets; columns = models ordered by parameter count. Pythia-1.4B's 100% accuracy on Biography (the green cell) suggests memorization.*